\documentclass{article} 
\usepackage[preprint]{colm2026_conference}

\usepackage{microtype}
\usepackage{hyperref}
\usepackage{url}
\usepackage{booktabs}

\usepackage{amsmath}
\usepackage{algorithm}
\usepackage{algorithmic}

\usepackage{multirow}

\usepackage{graphicx}

\usepackage{wrapfig}

\usepackage[table]{xcolor}

\usepackage{lineno}

\definecolor{darkblue}{rgb}{0, 0, 0.5}
\hypersetup{colorlinks=true, citecolor=darkblue, linkcolor=darkblue, urlcolor=darkblue}

\title{Mask Is What DLLM Needs: A Masked Data Training Paradigm for Diffusion LLMs}

\author{Linrui Ma, Yufei Cui\thanks{Project Leader}~,~Kai Han\thanks{Corresponding Authors}~\&~Yunhe Wang\footnotemark[2] \\
Noah's Ark Lab, Huawei\\
Montreal, Canada \& Beijing, China \\
\texttt{linrui.ma@\{h-partners.com, umontreal.ca\}}, \\
\texttt{\{yufei.cui, kai.han, yunhe.wang\}@huawei.com} \\
}

\begin{document}

\ifcolmsubmission
\linenumbers
\fi

\maketitle

\begin{abstract}
Discrete diffusion models offer global context awareness and flexible parallel generation. However, uniform random noise schedulers in standard DLLM training overlook the highly non-uniform information density inherent in real-world sequences. This wastes optimization resources on low-density structural glues while leaving high-density logical pivot points severely under-optimized. To address this, we propose an Information Density Driven Smart Noise Scheduler. By extracting information-dense hubs and applying Complementary Priority Masking, our method decouples a single training instance into mutually reinforcing reasoning and syntax samples, forcing the model to master both logical deduction and foundational sequence structure. Experiments demonstrate that our approach improves average accuracy by ~4\% across four Code and Math reasoning benchmarks, significantly outperforming uniform baselines. Mechanistic analyses further reveal that probabilistic priority masking effectively mitigates contextual collapse during block diffusion training. Overall, this density-aware strategy efficiently unlocks the reasoning potential of diffusion language models at minimal annotation cost, emerging as a promising new masked data training paradigm for Diffusion LLMs. Our processed dataset can be found \href{https://huggingface.co/datasets/malr07/opc-sft-stage2-dense-extracted}{here}.
\end{abstract}

\section{Introduction}
\label{sec:intro}

\begin{figure}[!htb]
  \centering 
  \makebox[\textwidth][c]{%
    \includegraphics[width=1.05\linewidth]{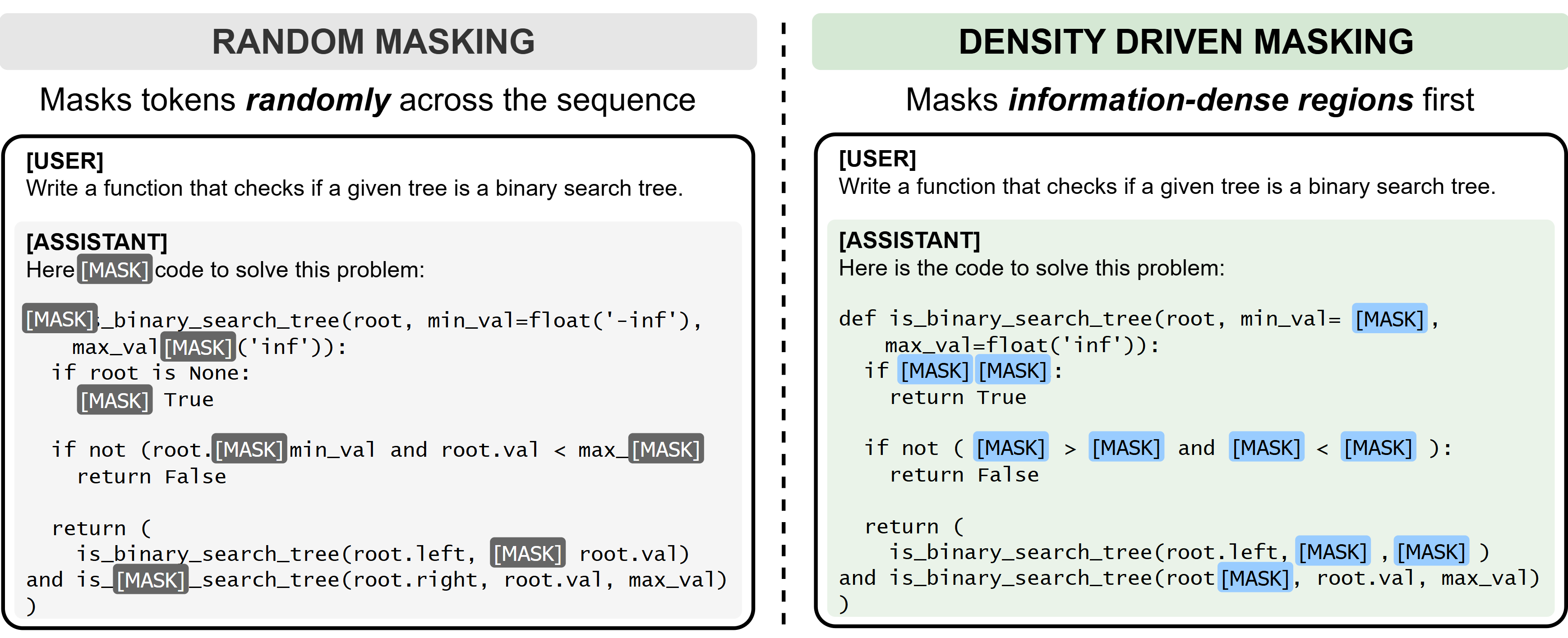}
  }
\caption{An illustrative comparison of random masking versus our density driven masking}
\label{fig:ill}
\end{figure}

Discrete Diffusion Models \citep{li2022diffusion, austin2021structured, nie2025largelanguagediffusionmodels} are gradually standing out as a powerful competitor against the Autoregressive (AR) language model architectures. Unlike the strict left-to-right word-by-word generation of AR models \citep{brown2020language, touvron2023llamaopenefficientfoundation}, the diffusion paradigm possesses highly flexible any-order parallel decoding capabilities. Furthermore, it exhibits better global context awareness and long-range planning abilities during the iterative denoising process \citep{pmlr-v162-janner22a, israel2025planneddiffusion}. This gives diffusion language models a significant advantage in tasks requiring dense logical deduction and global holistic arrangements. Recent successes of large-scale diffusion models further verify the huge potential of this paradigm under the Scaling Law \citep{kaplan2020scalinglawsneurallanguage}.

However, existing discrete diffusion models generally adopt a Uniform Random Masking strategy during training (e.g., supervised fine-tuning, SFT). Specifically, the model usually relies on an input-agnostic Noise Scheduler to control the noise addition ratio at different timesteps, and fits the data distribution by optimizing the evidence lower bound (ELBO) \citep{austin2021structured}. This input-agnostic scheduling mechanism assigns equal noising and denoising probabilities to all tokens in the sequence, yet it neglects the inherent information density distribution differences in real-world texts \citep{joshi-etal-2020-spanbert}. For example, when solving math problems or writing code, a real human will notice that the logical information entropy contained in critical deductive pivot points or control flow statements is much higher than that of ordinary conjunctions or punctuations. Existing schemes impose undifferentiated reconstruction constraints on all positions, which easily leads the model to waste optimization resources on low information density areas (such as syntactic glue), while facing a sub-optimal dilemma at information-dense hubs like critical logical leaps that decide the success or failure of the task.

Aiming at the above limitations, we propose an Information Density Driven Smart Noise Scheduler for better Diffusion Model training. The core intuition of this method originates from the \textit{Cloze Test} in human cognitive learning \citep{devlin-etal-2019-bert} : efficient learning often focuses on reconstructing core concepts rather than recovering redundant conjunctions. Based on this, we introduce a structured prior for the noising forward process of the diffusion model: first, we precisely identify the information-dense hubs in the training data through predefined rules or LLMs \citep{openai2024gpt4technicalreport}; Then, during noise scheduling, we assign significantly higher masking priority to these high-entropy spans, forcing the model to learn deep reasoning correlations. At the same time, to avoid the model losing basic language coherence, we introduce a Complementary Sampling mechanism, decoupling the sample into complementary mask pairs focusing on the logical skeleton and focusing on grammatical details, ensuring the comprehensiveness of the training objective. Figure \ref{fig:ill} gives an illustrative example of our proposed method applied on a sample code data entry.

Extensive experimental results show that, across four mainstream code and math reasoning benchmarks, the model equipped with this smart density driven scheduler achieves an average performance improvement of 4\% compared to the baseline model with standard random scheduling, proving the effectiveness of this scheme in enhancing the reasoning ability of diffusion models.

Overall, our main contributions are summarized as follows:
\begin{enumerate}
    \item We introduce a novel \textbf{Masked Data Training Paradigm} for Diffusion LLMs. By introducing a token-level, information-density-driven noising mechanism, we break the bottleneck of traditional input-agnostic, uniform noise scheduling.
    \item We propose a complete pipeline featuring automated information-dense hub extraction and complementary priority masking, significantly improving the performance of diffusion models on complex reasoning tasks without changing the base model architecture. We also open-source the dataset \href{https://huggingface.co/datasets/malr07/opc-sft-stage2-dense-extracted}{here}. 
    \item We provide mechanistic insights into block diffusion training, demonstrating that our probabilistic scheduling effectively avoids contextual collapse and achieves optimal performance leaps with minimum data annotation cost.
\end{enumerate}

\section{Methodology}

\begin{figure}[htb]
  \centering 
  \makebox[\textwidth][c]{%
    \includegraphics[width=1.1\linewidth]{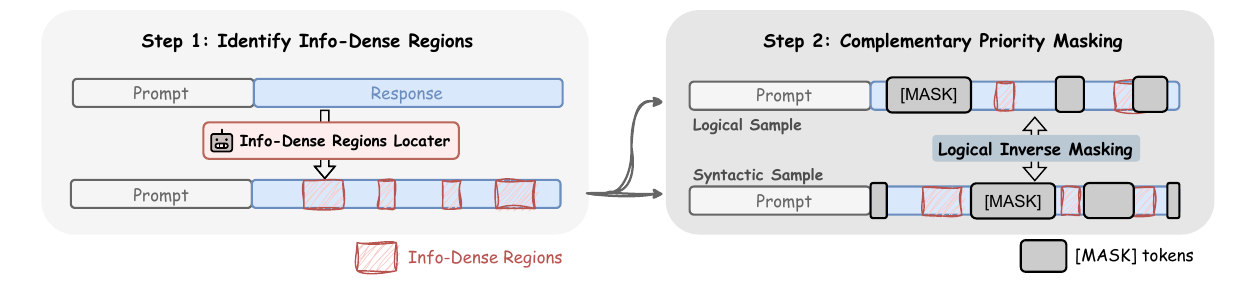}
  }
\caption{Overview of the Information Density Driven Noise Scheduler Pipeline for one sample. Left: The sample first undergoes LLM-based dense-region identification; Right: Then the sample is duplicated into two, and the \textit{Logical Sample} is made by prioritize masking dense-regions given noise ratio $\sigma_t$ and bias weight $w$, meanwhile the \textit{Syntactic Sample} is made by taking logical inverse of the Logical Sample's mask.}
\label{fig:overall}
\end{figure}

As shown in the figure \ref{fig:overall}, our proposed smart noise scheduling scheme contains two core stages: Information-Dense Region Extraction (left) and Complementary Priority Masking (right).

\subsection{Information-Dense Region Extraction}

To break the assumption of traditional discrete diffusion models on the uniform distribution of information density at each position of the text, we first need to conduct a fine-grained information density parsing on the training data. Given a target sequence $X = (x_1, x_2, \dots, x_N)$ with a length of $N$, our goal is to construct a corresponding binary indicator vector $C \in \{0, 1\}^N$, where $c_i = 1$ denotes that the $i$-th token belongs to the 'priority' region with high information density.

In the concrete implementation, we adopt an offline annotation pipeline \citep{alpaca} based on a strong language model. We input the instruction (Prompt) and the target answer (Answer) of the SFT stage together into the large language model, and prompt it to extract the core logic spans in the answer according to predefined structured rules:

\begin{itemize}
    \item For Code data: The info-dense hubs are defined as \textit{Control Flow Conditions} (i.e. judgment statements of \texttt{if}/\texttt{while}) and \textit{Algorithm Pivot Points}, etc.
    \item For Math data: The info-dense hubs are defined as \textit{Core Mathematical Operations} as well as \textit{Key Intermediate \& Final Results}, etc.
\end{itemize}

The extracted info-dense regions will be accurately mapped back to the original sequence, thereby generating the indicator vector $C$ as an additional property of this sample, which is used for subsequent noise scheduling. 

It is worth noting that this extraction module is designed to be Plug-and-play. In future work, we will explore hard rule matching based on Abstract Syntax Trees \citep{feng-etal-2020-codebert}, adaptive extraction based on the model's own confidence, and end-to-end learnable mask networks to further replace the current heuristic scheme.

\subsection{Complementary Priority Noise Scheduling}

After acquiring the info-dense region indicator vector $C$, we refine the noising distribution of the diffusion model during training.

\paragraph{Limitations of traditional uniform scheduling} In the standard discrete diffusion model training, given a randomly sampled timestep $t \sim \mathcal{U}(0, T)$ and a target mask rate $\sigma_t \in (0, 1)$ determined by the scheduler, the traditional scheme performs independent and identically distributed (I.I.D.) random masking for all positions \citep{Chang_2022_CVPR, austin2021structured}. Thus, for any position $i$, the probability of it being converted into a Mask Token is all:
$$P(m_i = 1) = \sigma_t$$
Among them, $M = (m_1, \dots, m_N) \in \{0, 1\}^N$ is the generated mask vector. This traditional approach prevents the model from performing targeted learning on critical reasoning breakpoints under the given timestep.

\paragraph{Priority Masking} 
To overcome the above problems, based on the indicator vector $C$ obtained earlier, we divide the sequence into priority regions ($c_i = 1$) and non-priority regions ($c_i = 0$). We set the probability of the priority region being masked as $w$ times of the non-priority region ($w > 1$). Let $p_{base}$ be the basic mask probability of the non-priority region, then the mask distribution is reconstructed as:
$$P(m_i = 1 | c_i) = \begin{cases} \min(w \cdot p_{base}, 1), & \text{if } c_i = 1 \\ p_{base}, & \text{if } c_i = 0 \end{cases}$$
Since we have introduced the spatial prior, to ensure that the overall marginal noise level of this sample still strictly obeys the $\sigma_t$ given by the scheduler, we dynamically calculate $p_{base}$ by solving the probability mass conservation equation:
$$\frac{1}{N} \sum_{i=1}^N P(m_i = 1 | c_i) = \sigma_t$$
We can therefore achieve information density focusing on the micro level while keeping the  noise schedule of the diffusion model unchanged on the macro level.

\paragraph{Complementary Masking \& Decoupling} 
\citet{clark2020electrapretrainingtextencoders, bie2025llada20scalingdiffusionlanguage} point out that only conducting model training on the masked tokens may lead to the ignorance of the unmasked (Clean) contexts. To maximize data utilization and increase training stability, we introduce the complementary masking mechanism. For the generated priority mask $M$, we construct its logical NOT mask $\bar{M}$ by taking $\bar{M} = \mathbf{1} - M$. In the same training mini-batch, the original sample $X$ is combined, respectively, with $M$ and $\bar{M}$ to generate two completely complementary training samples.

Specifically, when priority scheduling is combined with the complementary mechanism, an elegant density-based decoupling effect naturally emerges, the two complementary identical samples become one \textit{information-focused} sample and one \textit{structural-focused} sample:
\begin{itemize}
    \item \textbf{Information-focused Sample ($X_M$):} Since $M$ covers the logical key positions preferentially, this sample forces the model to cross a huge information gap and combine context in order to carry out the most difficult deep logical deduction step.
    \item \textbf{Structural-focused Sample ($X_{\bar{M}}$):} Under the complementary mask $\bar{M}$, those high information density regions are likely to be reserved as clean tokens, while the redundant positions, such as grammar glue and conjunctions, are covered. This encourages the model to prioritize learning of language structure coherence and fluency based on the logical skeleton.
\end{itemize}

This type of density-based decoupling provides high-level logical guidance and basic grammar consolidation during the training process, resulting in greater optimisation efficiency than conventional random complementary sampling.

\section{Experiments}

\subsection{Experimental Setup}

\paragraph{Model and Training Details} 
In all experiments, we adopt the LLaDA-2.0-mini model \citep{bie2025llada20scalingdiffusionlanguage} as the base model and use the dFactory framework. Regarding the hyper-parameter settings of the diffusion process, we fix the maximum sequence length as $N_{seq} = 2048$, and set the global batch size as 16. For block diffusion, the block size is set to 32, and the global noise rate $\sigma_t$ sampling interval generated by the scheduler is restricted in $[0.3, 0.8]$. In addition, we also regard the \texttt{<eos>} token as a maskable and predicted target to enhance the model's perception of sequence boundaries \citep{arriola2025block}. During testing, the maximum generation length is set to 512, and the decoding steps are consistent with training ($T=32$).

\subsection{Dataset Preparation} 
The fine-tuning (SFT) dataset comprises a mixture of OPC-SFT-Stage2 (code domain) \citep{Huang2024OpenCoderTO} and GSM8K (math domain) \citep{cobbe2021gsm8k}, totaling around 450K samples. The model is trained for one epoch on this data mixture.
We utilize GPT-4o \citep{openai2024gpt4technicalreport} as the offline feature extractor and extract the critical logic spans in the samples according to predefined prompts. For code or math datasets, we extract key regions from only a portion of the data (usually 10\% to 30\% of the entire dataset), leaving the remainder unchanged. The extracted portion is then shuffled with the unextracted remainder, ensuring the combined data remains consistent in content and total volume while enhancing its diversity. The entire processed OPC-SFT-Stage2 dataset can be find \href{https://huggingface.co/datasets/malr07/opc-sft-stage2-dense-extracted}{here}.

Letting the indicator vector $C \in \{0,1\}^N$ denote the extraction result, we calculated $\rho_{code}$ and $\rho_{math}$, the average per-sample proportion of high-density regions in the Code and Math datasets, respectively, and observed that $\rho_{code} \approx 0.25$ and $\rho_{math} \approx 0.31$, which provides a distributional prior for the subsequent priority noising process.

\subsection{Supervised Fine-Tuning Results}

\begin{table}[htbp]
\centering
\resizebox{\textwidth}{!}{
\begin{tabular}{llccccc}
\toprule
\multirow{2}{*}{Data} & \multirow{2}{*}{Method} & \multicolumn{2}{c}{Code Benchmarks} & \multicolumn{2}{c}{Math Benchmarks} & \multirow{2}{*}{AVG} \\
\cmidrule(lr){3-4} \cmidrule(lr){5-6}
& & HumanEval & MBPP & GSM8K & MATH500 & \\
\midrule
None & Original & 50.00 & 55.00 & \textbf{86.58} & 40.80 & 58.10 \\
Code + Math & Baseline & 57.93 & \textbf{56.80} & 69.14 & 37.40 & 55.32 \\
\rowcolor{gray!20} 
Code 10\% + Math 50\% & \textbf{Ours} ($w=2$) & \textbf{65.24} & 54.00 & 73.92 & \textbf{43.60} & \textbf{59.19} \\
\bottomrule
\end{tabular}%
}
\caption{Comparison of the original model, trained baseline, and our proposed density guide scheduling.}
\label{tab:main_results}
\end{table}

Table \ref{tab:main_results} displays the Pass@1 or Accuracy performance of different training strategies on four key Benchmarks (HumanEval \citep{chen2021evaluatinglargelanguagemodels}, MBPP \citep{austin2021programsynthesislargelanguage}, GSM8K \citep{cobbe2021trainingverifierssolvemath}, MATH500 \citep{lightman2024lets}). Among them, \textit{Original} represents the original off-the-shelf model without doing training; \textit{Baseline} represents the model trained with the same 450K mixed samples but with traditional uniform random masking (namely $w=1$); \textit{Ours} then represents the model adopting our proposed complementary priority masking scheduling (setting Code extraction ratio 10\%, Math extraction ratio 50\%, and bias weight $w=2$).

We draw two key conclusions from the above results:

\textbf{SFT Domain Shift:} As the SFT mixed data focuses strongly on code tasks and does not introduce alignment punishment mechanisms such as RLHF, the performance of all trained models on the code benchmarks is greatly improved compared with \textit{Original}. However, a certain degree of forgetting appears on the math benchmarks. This aligns with our expectations at the SFT stage.

\textbf{Superiority of Smart Scheduling:} Given identical training data and computational overhead, our proposed smart scheduler (\textit{Ours}) achieves a comprehensive superiority over the \textit{Baseline}. It performs exceptionally well on code generation (+7.31\% on HumanEval) and complex mathematical reasoning (+6.20\% on MATH500), leading the average score (AVG) across all four tasks to increase significantly from 55.32 to 59.19 (+3.87\%). This powerfully demonstrates that density-aware mask scheduling can significantly improve both optimization efficiency and generalization capacity when training data is limited.

\subsection{Ablation Studies}

\subsubsection{Impact of Bias Weight $w$}


\begin{figure}[htb]
  \centering
  \makebox[\textwidth][c]{%
    \includegraphics[width=0.75\linewidth]{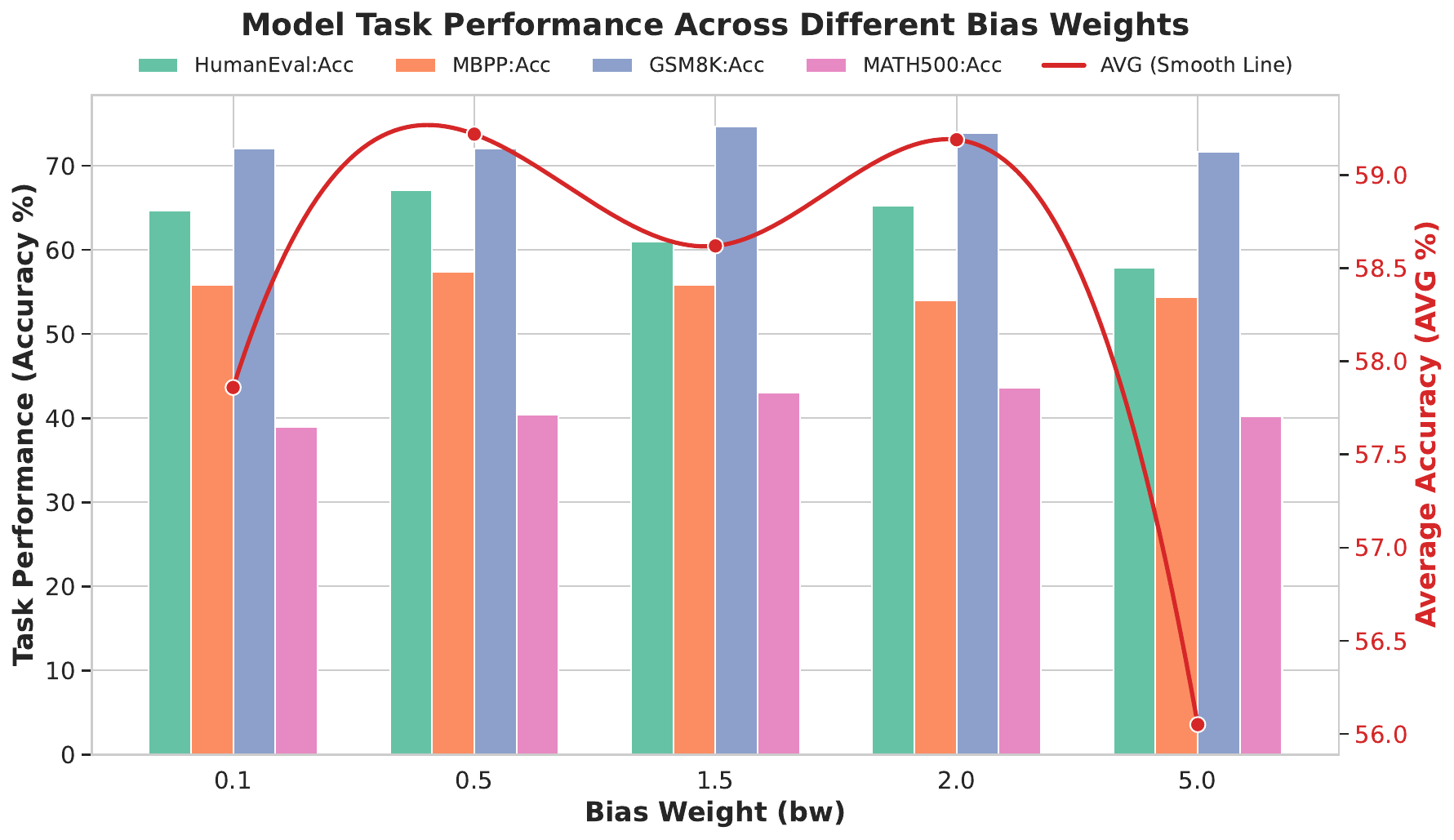}
  }
\caption{Impact of different bias weights ($w$) on fixed preprocessed data (Code 10\% + Math 50\%)}
\label{fig:bias weight}
\end{figure}

The bias weight $w$ decides the intensity of the mask distribution tilting towards high information density areas. When $w=1$, the strategy degenerates into traditional random masking (Baseline). We tested the performance of the model when $w \in \{0.1, 0.5, 1.5, 2, 5\}$, figure \ref{fig:bias weight} shows the result.

The results reveal a distinctly symmetric distribution of average scores. Since we jointly optimize the priority mask $M$ and its logically complementary mask $\bar{M} = \mathbf{1} - M$ within the same minibatch, the training data distributions induced by a bias weight $w$ and its reciprocal $1/w$ are expected to be isomorphic in terms of mathematical expectation. The empirical data perfectly corroborates this deduction: the model performances at $w=2$ (AVG: 59.19) and $w=0.5$ (AVG: 59.22) are highly consistent, with both yielding the optimal training outcomes.

The results also show that over-constraint can harm the model. Weak priors (such as $w = 1.5$) typically only offer limited improvement to the model's capabilities. Conversely, extremely strong priors (such as $w = 5$ or $w = 0.1$) negatively impact model performance, reducing the average score to approximately 56.05. This suggests that excessive distortion of the distribution will prevent the joint probability language modeling by the diffusion model, and that a moderate density-driven approach ($w\approx2$) is optimal for overcoming the optimization bottleneck.

\subsubsection{Soft vs. Hard Priority Masking}

\begin{table}[htbp]
\centering
\resizebox{\textwidth}{!}{
\begin{tabular}{llccccc}
\toprule
\multirow{2}{*}{Code Processed} & \multirow{2}{*}{Method} & \multicolumn{2}{c}{Code Benchmarks} & \multicolumn{2}{c}{Math Benchmarks} & \multirow{2}{*}{AVG} \\
\cmidrule(lr){3-4} \cmidrule(lr){5-6}
& & HumanEval & MBPP & GSM8K & MATH500 & \\
\midrule
\multirow{2}{*}{10\%} & Hard Sample & 62.20 & 54.80 & 72.40 & 40.00 & 57.35 \\
& Soft Priority ($w=2$) & \textbf{64.02} & \textbf{56.20} & \textbf{73.39} & \textbf{44.20} & \textbf{59.45} \\
\midrule
\multirow{2}{*}{20\%} & Hard Sample & 53.05 & \textbf{50.00} & \textbf{73.84} & 34.40 & 52.82 \\
& Soft Priority ($w=2$) & \textbf{62.20} & 49.60 & 70.36 & \textbf{44.00} & \textbf{56.54} \\
\midrule
\multirow{2}{*}{30\%} & Hard Sample & 61.59 & \textbf{53.60} & 71.49 & 40.20 & 56.72 \\
& Soft Priority ($w=2$) & \textbf{63.41} & 50.20 & \textbf{75.66} & \textbf{43.60} & \textbf{58.22} \\
\midrule
\multirow{2}{*}{100\%} & Hard Sample & - & - & - & - & - \\
& Soft Priority ($w=2$) & 67.07 & 53.80 & 72.64 & 40.40 & 58.48 \\
\bottomrule
\end{tabular}
}
\caption{Comparison of Hard vs. Soft Priority Masking across different data scaling ratios.}
\label{tab:ablation_hard_soft}
\end{table}

In order to further explore the noising mechanism, we designed the extreme baseline of \textit{Hard Sample} (namely $w \to \infty$).  Under this setting, the scheduler will exhaustively preferentially cover all the key areas ($c_i=1$) until $\sigma_t$ is used up, and then cover the non-key areas.

As shown in table \ref{tab:ablation_hard_soft}, under the same proportion of data, the performance of the hard sample falls far behind that of the soft sample ($w=2$) and is even lower than the baseline. We attribute this to the \textit{contextual collapse} effect unique to block diffusion. Since the extracted mathematical or code info-dense regions are often continuous in physical space, deterministic hard masking will create large, continuous \textit{information black holes} in the text at a medium noise ratio ($\sigma_t \approx 0.5$). Without the local anchors, the gradient trajectory of block generation becomes too steep, resulting in training collapse. Conversely, our proposed Soft Priority Mask retains probabilistic local prompts, making the ELBO optimization smoother.

\subsubsection{Data Scaling Effect}

We evaluated the cost-effectiveness of conducting offline API extraction by applying our extraction pipeline on 10\%, 20\%, and 30\% of the Code data in Table \ref{tab:ablation_hard_soft} (note that the remaining 90\%, 80\%, or 70\% of the data are kept unprocessed to ensure the total number of samples is identical). We also applied extraction to 100\% of the code data to validate the model performance under an extreme data prior ratio.

The results demonstrate that our method achieves exceptional data efficiency: applying the information-density priority mask to only 10\% of the training data is sufficient to drive a substantial performance leap (AVG from 55.32 to 59.45).  Interestingly, as the preprocessing ratio scales up to 30\% and eventually reaches the extreme of 100\%, the overall performance improvement exhibits a clear saturation trend.

It is particularly worth noting that under the 100\% setting, while the domain-specific metric reaches its absolute peak (HumanEval: 67.07), the complex mathematical reasoning performance suffers a certain degree of degradation (MATH500 drops from the peak of 44.20 down to 40.40). We attribute this phenomenon to the aggravation of the domain shift: excessive structural priors injected into the Code domain crowd out the model's generalization capacity on other reasoning tasks.

Consequently, this discovery powerfully underscores the highly lightweight nature of our approach for practical deployment. Full-scale (100\%) data annotation is not only unnecessary but potentially sub-optimal for holistic reasoning. By incurring only a marginal cost for advanced API annotations on a tiny fraction of data (e.g., 10\%), developers can significantly push the overall reasoning upper limit of the base diffusion model.

\subsubsection{Without Complementary Masking}

\begin{figure}[htb]
  \centering 
  \makebox[\textwidth][c]{%
    \includegraphics[width=0.75\linewidth]{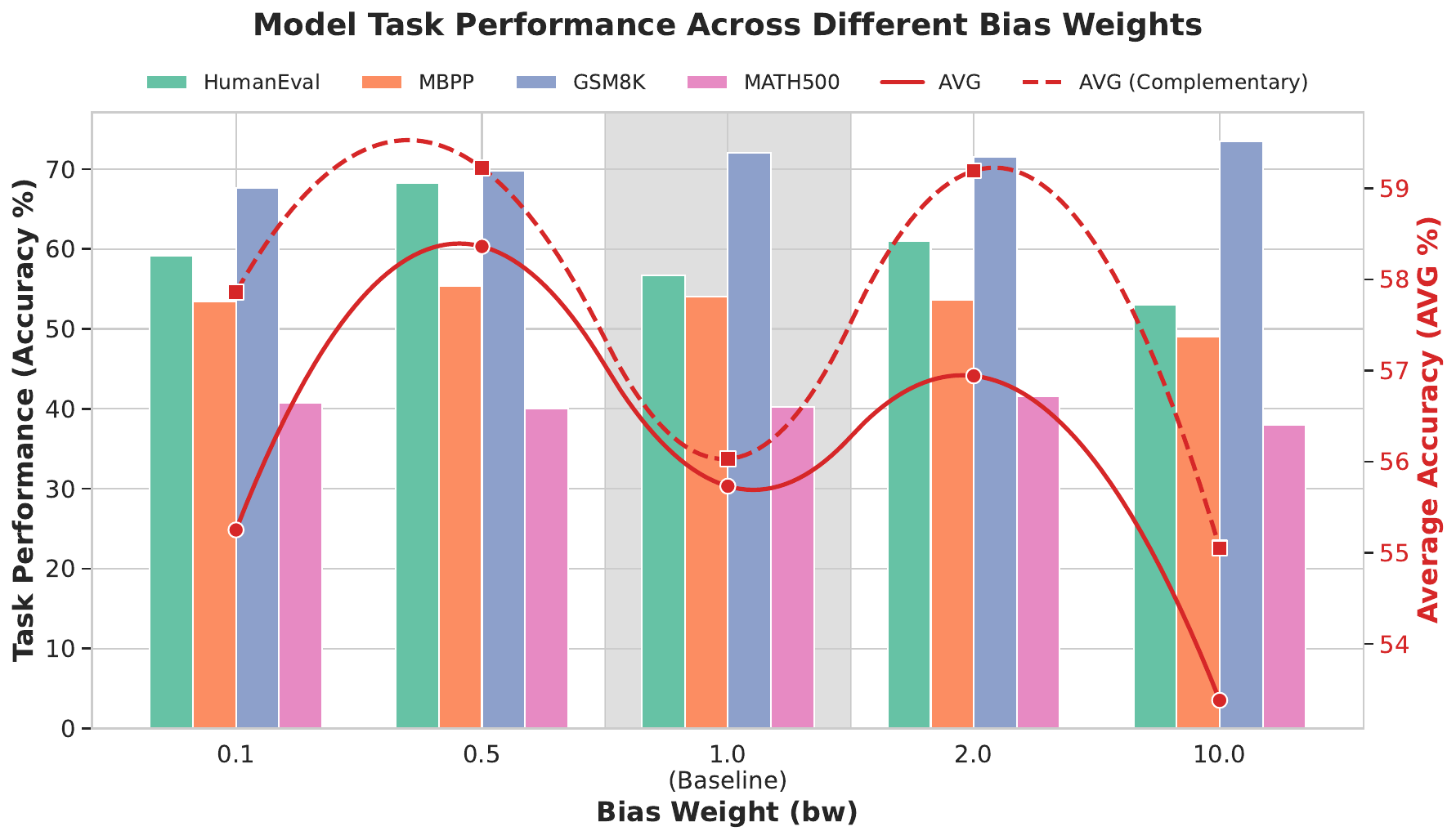}
  }
\caption{Performance of models trained without complementary masking (priority masking only). Plot area with gray background indicates baseline performance with bias weight $w$ set to 1. The red dashed line indicates the results with complementary sample under the same settings.}
\label{fig:abla}
\end{figure}

To further isolate the contribution of complementary masking, we conducted an ablation study using only priority masking (without the logical NOT counterpart $\bar{M}$). As shown in Figure \ref{fig:abla}, we evaluate model performance across different bias weights $w \in \{0.1, 0.5, 1.0, 2.0, 10.0\}$.

From the results, we can draw the following critical observations:

\textbf{Breaking of Distribution Symmetry:} Unlike the full method, the distribution symmetry is broken here. Specifically, $w=0.5$ (AVG $\approx 58.5\%$) significantly outperforms $w=2.0$ (AVG $\approx 57.0\%$). We attribute this to the fact that when the model only has a single training view, reserving high-entropy logic skeletons as visible anchors ($w=0.5$) provides a much more stable optimization trajectory than heavily masking them ($w=2.0$) without a learning buffer.

\textbf{Superiority of Information Density Priors:} Despite lacking complementary views, moderate semantic masking ($w=0.5$ and $w=2.0$) still successfully surpasses the uniform random baseline ($w=1.0$, gray area). Meanwhile, extreme priors (e.g., $w=10.0$) lead to severe performance degradation, verifying that too much distribution distortion destroys the ELBO optimization.

Most importantly, the peak performance achieved here still falls short of our full Complementary Priority Masking approach (red dashed line). This explicitly proves that the Density-Based Decoupling effect is an indispensable component for maximizing the reasoning potential of diffusion language models.

\section{Conclusion}

In this report, we identified a critical bottleneck in the Supervised Fine-Tuning of discrete diffusion models: the blindness of uniform random masking to the highly non-uniform information density of real-world sequences. To address this, we introduced an Information-Density-Aware Smart Noise Scheduler, replacing input-agnostic noise injection with a structurally informed, priority-driven masking paradigm. 

By extracting high-density structural pivots via an offline pipeline and applying Complementary Priority Masking, we elegantly decoupled the diffusion denoising objective into two distinct learning trajectories: deep logical deduction and fluent syntactic structuring.  Empirical results validate that this approach significantly enhances zero-shot reasoning capabilities on complex Code and Math tasks (+4\% average improvement) while demonstrating exceptional data efficiency. 

Furthermore, our ablation studies yielded critical insights into Block Diffusion dynamics. We mathematically and empirically demonstrated the symmetry of complementary bias weights ($w$ and $1/w$), and revealed that probabilistic \textit{Soft Priority Masking} effectively avoids the fatal contextual collapse inherently caused by deterministic \textit{Hard Masking}.

\textbf{Future Work:} While the current LLM-based extraction serves as a highly effective proof-of-concept, it introduces dependency on external language models. Future work will evolve this pipeline into a fully self-contained ecosystem. Promising directions include exploring rule-based extraction via Abstract Syntax Trees (AST) for strict programmatic domains, or designing an end-to-end learnable masking module that dynamically discovers high-density hubs based on the diffusion model's own real-time loss landscape and confidence metrics.

\newpage

\bibliography{colm2026_conference}
\bibliographystyle{colm2026_conference}


\end{document}